\documentclass[letterpaper, 10 pt, journal, twoside]{IEEEtran}

\IEEEoverridecommandlockouts                              

\usepackage{graphics} 
\usepackage{epsfig} 
\usepackage{mathptmx} 
\usepackage{times} 
\usepackage{amsmath} 
\usepackage{amssymb}  
\usepackage{balance}
\usepackage{siunitx}
\usepackage{comment}
\usepackage{xcolor}
\usepackage{dirtytalk}
\usepackage{gensymb}

\newcommand\revision[1]{\textcolor{black}{#1}}

\begin{document}

\title{The Folded Pneumatic Artificial Muscle (foldPAM): Towards Programmability and Control via End Geometry}

\author{Sicheng Wang$^{1}$, Eugenio Frias Miranda$^{1}$, and Laura H. Blumenschein$^{1}$
\thanks{Manuscript received: August 24, 2022; Revised: December 1, 2022; Accepted: January 7, 2023.}
\thanks{This paper was recommended for publication by Editor Yong-Lae Park upon evaluation of the Associate Editor and Reviewers' comments.}
\thanks{$^{1}$School of Mechanical Engineering, Purdue University, West Lafayette, IN 47906, USA}
\thanks{Email: \tt\footnotesize \{wang5239, efrias, lhblumen\}@purdue.edu }
\thanks{All funding for this project was provided by Purdue University.}
\thanks{Digital Object Identifier (DOI): 10.1109/LRA.2023.3238160}}

\maketitle

\markboth{IEEE Robotics and Automation Letters. Preprint Version. Accepted January, 2023}
{Wang \MakeLowercase{\textit{et al.}}: The Folded Pneumatic Artificial Muscle} 

\begin{abstract}
 Soft pneumatic actuators have seen applications in many soft robotic systems, and their pressure-driven nature presents unique challenges and opportunities for controlling their motion. In this work, we present a new concept: designing and controlling pneumatic actuators via end geometry. We demonstrate a novel actuator class, named the folded Pneumatic Artificial Muscle (foldPAM), which features a thin-filmed air pouch that is symmetrically folded on each side. Varying the folded portion of the actuator changes the end constraints and, hence, the force-strain relationships. We investigated this change experimentally by measuring the force-strain relationship of individual foldPAM units with various lengths and amounts of folding. In addition to static-geometry units, an actuated foldPAM device was designed to produce continuous, on-demand adjustment of the end geometry, enabling closed-loop position control while maintaining constant pressure. Experiments with the device indicate that geometry control allows access to different areas on the force-strain plane and that closed-loop geometry control can achieve errors within 0.5\% of the actuation range.
\end{abstract}


\section{Introduction}
\IEEEPARstart{T}{he design} of pneumatic artificial muscles (PAMs) remains an active field of research despite dating back to the early days of robotics \cite{baldwin69}. This actuator type is commonly seen in soft robots \cite{rus15}, and has been used in a range of applications from haptics \cite{zhu20} to prosthetics \cite{andrikopoulos11}. Typically, a PAM has only a single degree of freedom but that degree of freedom can produce contraction, expansion, bending, etc. upon the addition or removal of air~\cite{hawkes16,li17,tawk18}. 

Existing PAM designs can be classified based on their material and principle of motion. With a focus on materials, we break the designs into categories of “elastomer-based” and “thin-filmed” actuators.  Elastomer-based PAM designs feature partially constrained flexible materials that expand or contract elastically upon change in pressure. McKibben Pneumatic Muscle \cite{baldwin69} and the Pleated Pneumatic Artificial Muscle \cite{daerden01} are examples that have a single air bladder and flexible, enveloping constraints. PneuNet actuators \cite{ilievski11}, in contrast, have networks of multiple interconnected, elastic air chambers constrained by material thickness or embedded fibers. The fabric PAM \cite{naclerio20} further develops the concept by synthesizing the functions of the bladder and the constraint into a single material: highly compliant, air-tight, bias-cut fabric. 

Thin-filmed PAMs, on the other hand, feature a thin, nearly inelastic but flexible film fabricated into a pouch. Examples within this class of actuators are often relatively simple to manufacture, involving procedures such as heat stamping, as for Peano/AeroMorph actuators \cite{ou16}, or attaching rigid diametric constraints on a thin-film tubing segment, as in Serial Pneumatic Artificial Muscles (sPAM) \cite{greer17} and Bubble Artificial Muscles (BAM) \cite{diteesawat15}. Both linear and rotational motions may be achieved and the concept has also inspired the design of actuators based on other forms of energy and materials, such as the HASEL-Peano actuator \cite{kellaris18}.

\begin{figure}[tb]
    \centering
    \includegraphics[width=.95\columnwidth]{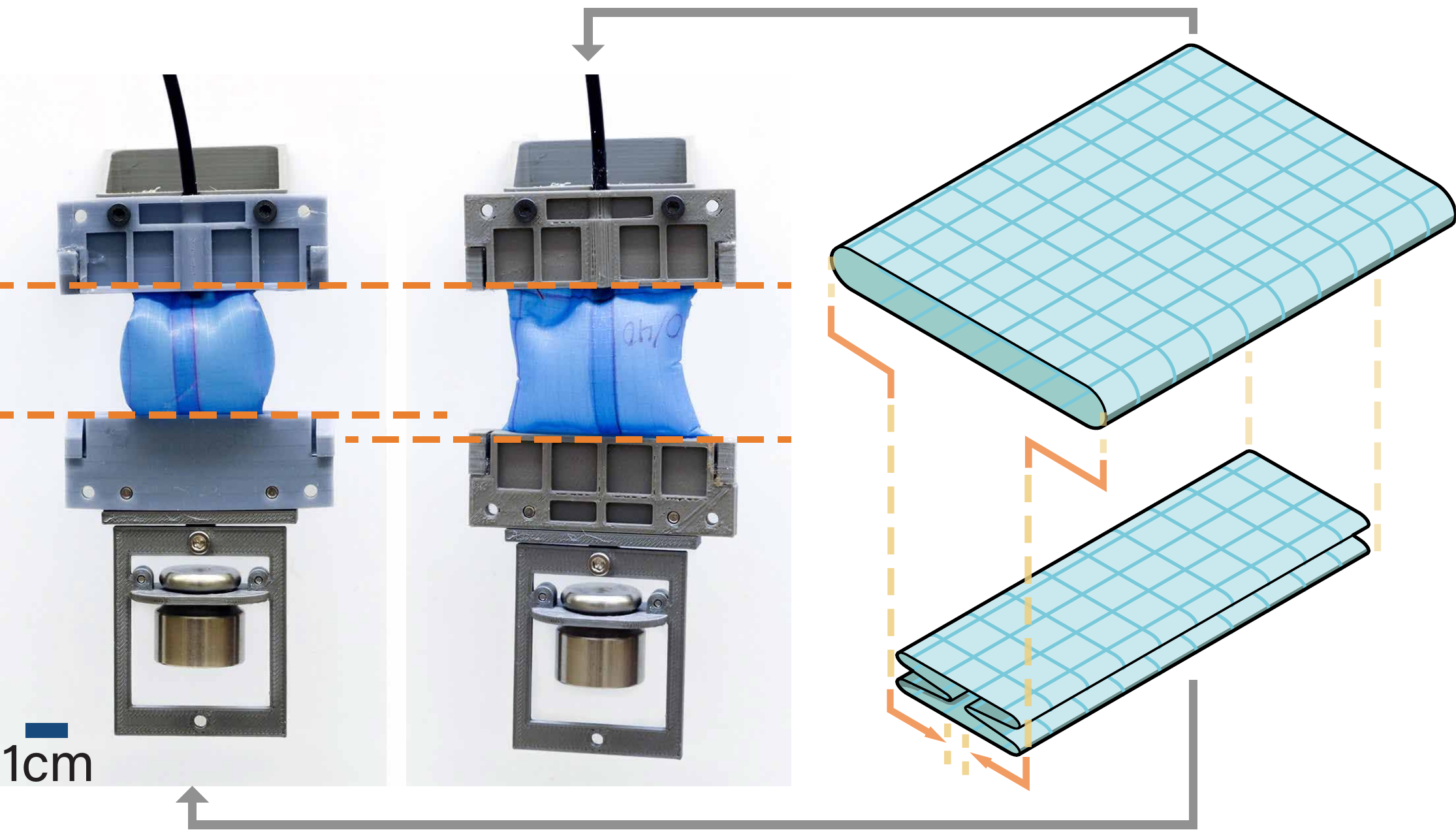}
    \caption{The conceptual illustration of the foldPAM, showing its cross-section and different strains produced by foldPAM units with a fold ratio ($f_r$) of 0.4 (left) and 0 (right) respectively, when subjected to a load of approximately 50g mass.}
    \label{fig:concepts}
    \vspace{-1.5em}
\end{figure}

For most of the PAM designs, the force and displacement produced by an actuator at a given input pressure is a functional relationship determined by its physical construction. With emerging applications that demand precise control over both force and displacement, the programmability of PAMs has received increased attention. Methods to design and fabricate actuators to match desired trajectories and ranges of motion have been developed, especially for elastomer-based actuators as its structure enables a large design space with many parameters: actuators capable of producing a range of curvatures and torsions can be programmed by varying the orientation of constraining fibers and material stiffness \cite{connolly17}, selectively applying elastic and inelastic fabrics as constraint on a McKibben-type actuator \cite{luo22}, and varying the relative position of air channels connecting consecutive air chambers of a PneuNet actuator \cite{kan22}. A mechanism for actively modifying the behavior of an actuator by varying enclosure fiber angle was developed by Yoshida et al.~\cite{yoshida20}.

For the thin-filmed PAMs, the programmability at a single-unit scale can be achieved by varying the geometry, for example the dimension of a heat-sealed pattern at the middle of a Pouch Motor \cite{ou16}. While a number of geometric parameters have been shown to affect thin-filmed PAMs behavior, e.g. the aspect ratio \cite{veale16} and end radius \cite{diteesawat15}, such effects are more often considered from an optimization point of view, where maximizing metrics such as zero-tension strain and maximum force output is the primary concern. \revision{In this paper, instead of seeking a higher maximum force or strain, we take a close look at the potential to program or control a pneumatic actuator with geometry, without loss of performance compared with existing actuators.}

\revision{As a result,} we propose a new concept for a programmable thin-filmed PAM, the folded Pneumatic Artificial Muscle (foldPAM), constructed by a short segment of inextensible thin-film tubing that is folded symmetrically with respect to its principal axis. Adjusting the amount of folding results in a tunable force-strain relationship, allowing behaviors ranging from Pouch-Motor-like \cite{niiyama15} to a limit of sPAM-like \cite{greer17}. The force-strain relationship can be pre-configured or adjusted on-demand when an additional actuated degree of freedom is added. Actuating the folding further enables regulating the foldPAM output while at a constant input pressure.

In the following sections, we first discuss the notion of altering the actuator behavior by changing its end geometry. We then describe the design and fabrication of the foldPAM with static geometry, followed by experimental results for the force-strain curves produced by a foldPAM, compared to existing modeling results, with various design parameters. This experimentation produces a ``design space" as a collection of all possible force-strain curves. We then present the design and characterization of a foldPAM with active, reconfigurable geometry (``Active foldPAM"). Finally, we demonstrate the Active foldPAM controlling its output position under a step-change in external load using end geometry.

\section{Design and Control via End Geometry}
\revision{As noted in several previous studies, the inflated geometry, and hence the force-displacement relationship, of a serial Pneumatic Artificial Muscle (sPAM) or Bubble Artificial Muscle (BAM) is affected by the diameter of the constriction at the ends \cite{greer17,diteesawat15}. Drawing inspiration from these results, we hypothesize that for other thin-filmed actuators the force-strain relationship may also be modified by manipulating the end constraints -- in other words, we can utilize end geometry to pre-program or control in real-time the behavior of an actuator, which, for example, allows the actuator to achieve desired behaviors with a constant pressure source. Especially with the ability to modify this end geometry on-demand, the resulting actuator is analogous to a reducer with continually variable transmission ratio that regulates output of a system without significantly changing the operating point of the motor (compressor). For soft pneumatic actuators, this is a new degree-of-freedom that has not been explored before.}

\revision{To implement this strategy, we need to find a parameterized geometry such that a range of actuator performance can be obtained by varying one or more geometric parameters. The previous examples of sPAM and BAM suggest varying the diameter of the circular constriction at the ends, i.e. the circumference of the ends. However, maintaining a circular end introduces the challenge of creating a variable diameter device. Still, other geometries may be utilized to produce a variable circumference. In this study, we propose a realization of the variable end circumference by varying a symmetrical folding of the thin-film material at each end (Figure~\ref{fig:fopam_paras}(b)-(c)). With zero folding, the actuator forms a simple rectangular pouch; as the amount of folded material increases, the end width of the actuator reduces; at maximum folding, the end width of the actuator reduces to a third of its original width, and the folds fully overlap.}

\revision{While we use the folded geometry here to demonstrate the programmability enabled by end geometries, other parameterizable shapes, for example circular ends with variable diameters, are also viable, given that the design problems are solved. Using other end geometries may provide different design spaces to choose actuator behaviors, and the investigation of this space is left for future work.}

\section{Static Geometry foldPAM}
\subsection{Design and Fabrication}
\begin{figure}[tb]
    \centering
    \includegraphics[width=.9\columnwidth]{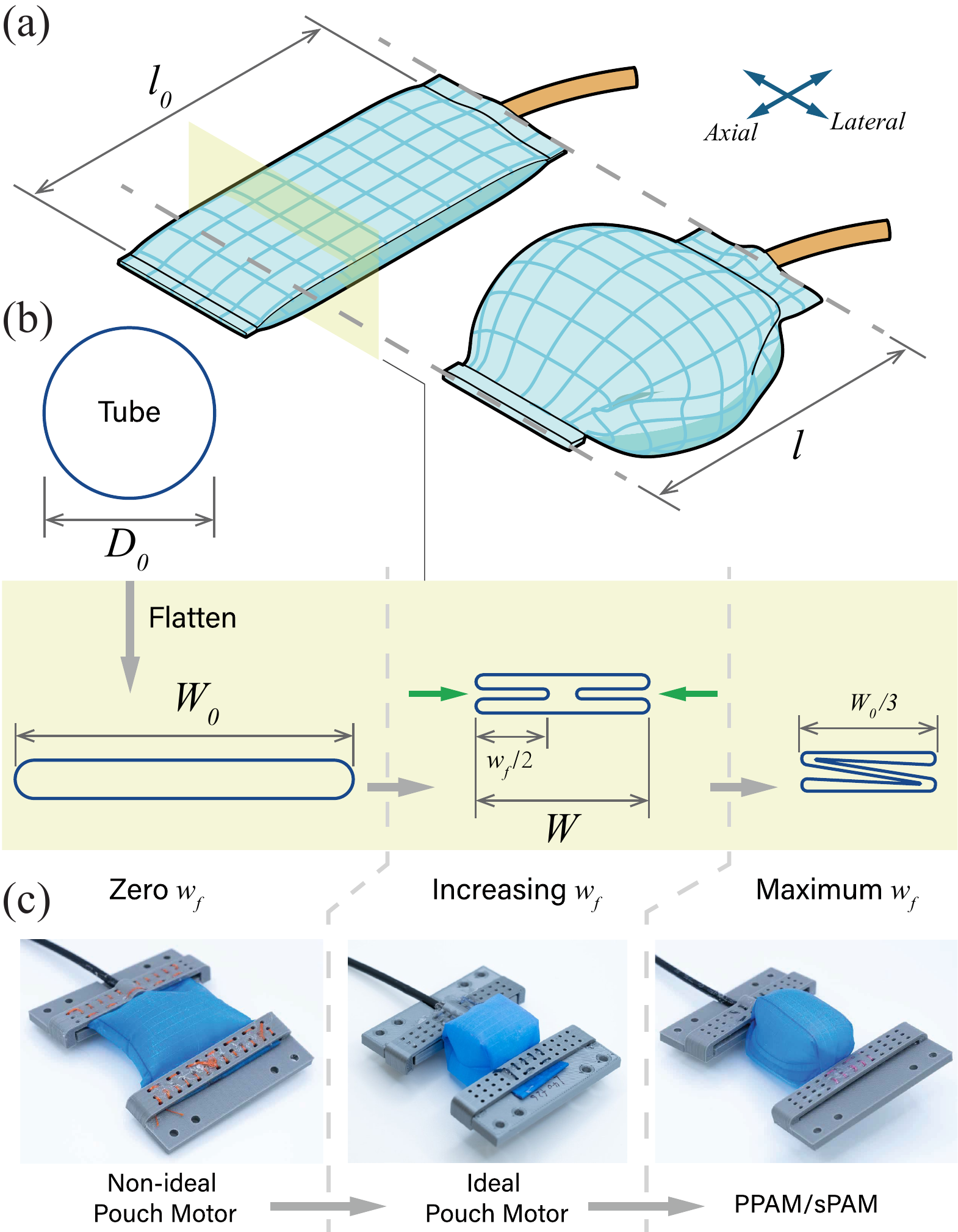}
    \caption{\revision{(a): A conceptual illustration of the foldPAM at deflated and inflated states. (b): The end cross-section with the folded width $w_f$ varying from 0 to its maximum value of $W_0/3$. (c): The foldPAM unit prototype at various folding and the analogous type of existing actuators, from zero to maximum folding: non-ideal pouch motor, ideal pouch motor, and sPAM/PPAM.}}
    \label{fig:fopam_paras}
    \vspace{-1em}
\end{figure}
\label{stat_fab}
In describing the design and fabrication process of a static foldPAM unit, we first introduce the parameters used to define a foldPAM actuator. The relevant variables are illustrated in Figure~\ref{fig:fopam_paras}. The actuator is made of a section of thin film tubing flattened to a rectangular piece of length $l_0$ and width $W_0 ={\pi}{D_0}/2$, where $D_0$ is the inflated diameter of the tubing. Each of the edges along the axial direction is then folded laterally in between the surfaces of the tubing by a distance of $w_f/2$, where $w_f$ is the total folded width on each side, resulting in an actuator with uninflated width of $W=W_0-w_f$. As a result, the geometry of a foldPAM unit can be described by two normalized quantities, fold ratio $f_r$ and aspect ratio $a_r$, defined by the following equations:
\begin{equation}
        f_r = \frac{w_f}{W_0},\;a_r = \frac{l_0}{W_0}.
\end{equation}

\begin{figure}[t]
    \centering
    \includegraphics[width=.95\columnwidth]{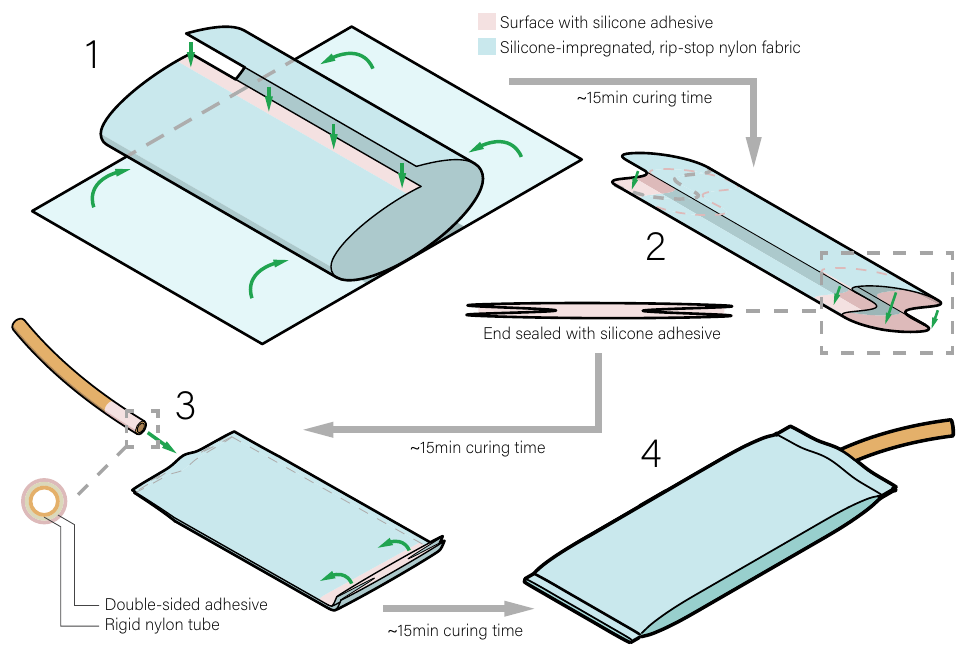}
    \caption{\revision{Fabrication process of a static foldPAM unit. 1: Roll a flat sheet of air-tight nylon fabric into a short segment of tube and make a lap joint with adhesive. 2: apply silicone adhesive to parts of the inner and outer surface, then seal one end of the unit. 3: Insert a thin tube for air supply then seal the other side of the unit. A double-sided adhesive (Red-e Tape, True Tape LLC.) compatible with both the tube material and the silicone adhesive is applied as the medium between the two. The end without air supply tube is folded and glued to ensure air-tightness. 4: finished actuator prototype. The above procedures may be performed in a different order.}}
    \label{fig:fopam_fab}
    \vspace{-1.5em}
\end{figure}
A foldPAM unit may be constructed with virtually any thin film that is flexible, near non-stretchable, and airtight. In this work we fabricated foldPAMs with 70$\mu{m}$ thick, silicone-impregnated, rip-stop nylon fabric (Seattle Fabrics, WA) and the ends were sealed with a silicone adhesive (Sil-Poxy, Smooth-On Inc., PA). Figure~\ref{fig:fopam_fab} shows the fabrication process. \revision{Additionally, a demonstration of the fabrication process can be found in the supplementary media file of the paper.} \revision{FoldPAMs can also be fabricated with other thin film materials, such as low-density polyethylene (LDPE) film. These materials obtained similar performance when tested with the method described in Section~\ref{sec:static_experiment}. We chose rip-stop nylon for the following experiments and demonstrations for its robustness and high strength. The material is estimated capable of sustaining an internal pressure on the order of 100kPa based on experimentally obtained tensile failure point, sufficient for our experiments. Further investigation is needed to determine the material and manufacturing procedure that delivers best performance for a specific application.}

\subsection{Workspace Prediction with Existing Actuator Concepts}
\label{sec:static_model}
For a given aspect ratio $a_r$, as the fold ratio $f_r$ varies from 0 to 0.67, the foldPAM can be seen as morphing through three states that are similar to existing actuator designs. As shown in Figure~\ref{fig:fopam_paras}(c), at $f_r = 0$, a foldPAM behaves like a non-ideal pouch motor \cite{niiyama15} with a length of $l_0$ and a width of $W_0$, since it is subjected to a loss of maximum force and strain due to its finite width. Currently, there lacks an analytical model that describes such boundary effect, though works such as Veale et al. \cite{veale16} have investigated the effect of pouch geometry experimentally.  As $w_f$ increases, the behavior of the actuator approaches an ideal pouch motor, as the folded portion provides additional material to compensate for the boundary effect. When the folded portion is sufficient to allow a circular section for the inflated actuator, i.e.,
\begin{equation}
    w_{f,circ} = \frac{2l_0}{\pi},
    \label{eqn:wf_circ}
\end{equation}
the actuator becomes an ideal pouch motor, as in Figure~\ref{fig:fopam_paras}(c). The force $F$ and strain $\epsilon$ of the actuator, parameterized by $\theta\in(0,\frac{pi}{2}]$, is given in \cite{niiyama15} by
\begin{equation}
    F(\theta) = (W_0-w_{f,circ}){l_0}P\frac{\cos\theta}{\theta}\\
    \label{eqn:pm_f}
\end{equation}
\begin{equation}
    \epsilon(\theta) = 1-\frac{\sin\theta}{\theta},
    \label{eqn:pm_eps}
\end{equation}
where P is the pressure within the actuator, and the strain $\epsilon$ for an actuator contracted to a length of $l$ is defined as
\begin{equation}
    \epsilon = 1-\frac{l}{l_0}.
    \label{eqn:eps_def}
\end{equation}

When the actuator reaches the maximum $f_r$, as in Figure~\ref{fig:fopam_paras}(c), we approximate its force-strain behavior by the serial/Pleated PAM (sPAM/PPAM) model \cite{greer17}. We take the approximation because the actuator is highly constricted at the ends and has a large apparent aspect ratio ($l_0/W$) at high fold ratios, as similar to the two existing class of actuators. For a given contraction $\epsilon$, the force produced by the maximally folded foldPAM is given by
\begin{equation}
    F(\epsilon) = {\pi}Ph^2\frac{1-2m}{2m\cos^2\phi},
    \label{eqn_ppam_f}
\end{equation}
and $m$, $\phi$ are constants satisfying
\begin{equation}
    \begin{gathered}
        \frac{E(\phi{\backslash}m)}{\sqrt{m}\cos\phi} = \frac{l_0}{h}(1-\frac{\epsilon}{2})\\
        \frac{F(\phi{\backslash}m)}{\sqrt{m}\cos\phi} = \frac{l_0}{h},
    \end{gathered}
    \label{eqn:ppam_const}
\end{equation}
where $F(\phi{\backslash}m)$ and $E(\phi{\backslash}m)$ are incomplete elliptic integral of first and second kind respectively, and we use the overall thickness of the actuator $h$, as indicated in Figure~\ref{fig:fopam_paras} to replace the term for end constriction radius in the sPAM/PPAM model. In summary, while there is not an analytical model that describes the non-ideal pouch motor state, making it difficult to predict the no-folding limit of the foldPAM performance, We hypothesize that the pouch motor and sPAM/PPAM models predict the location of the force-strain curves when the foldPAM attains these states.

\begin{figure}[t]
    \centering
    \includegraphics[width=.95\columnwidth]{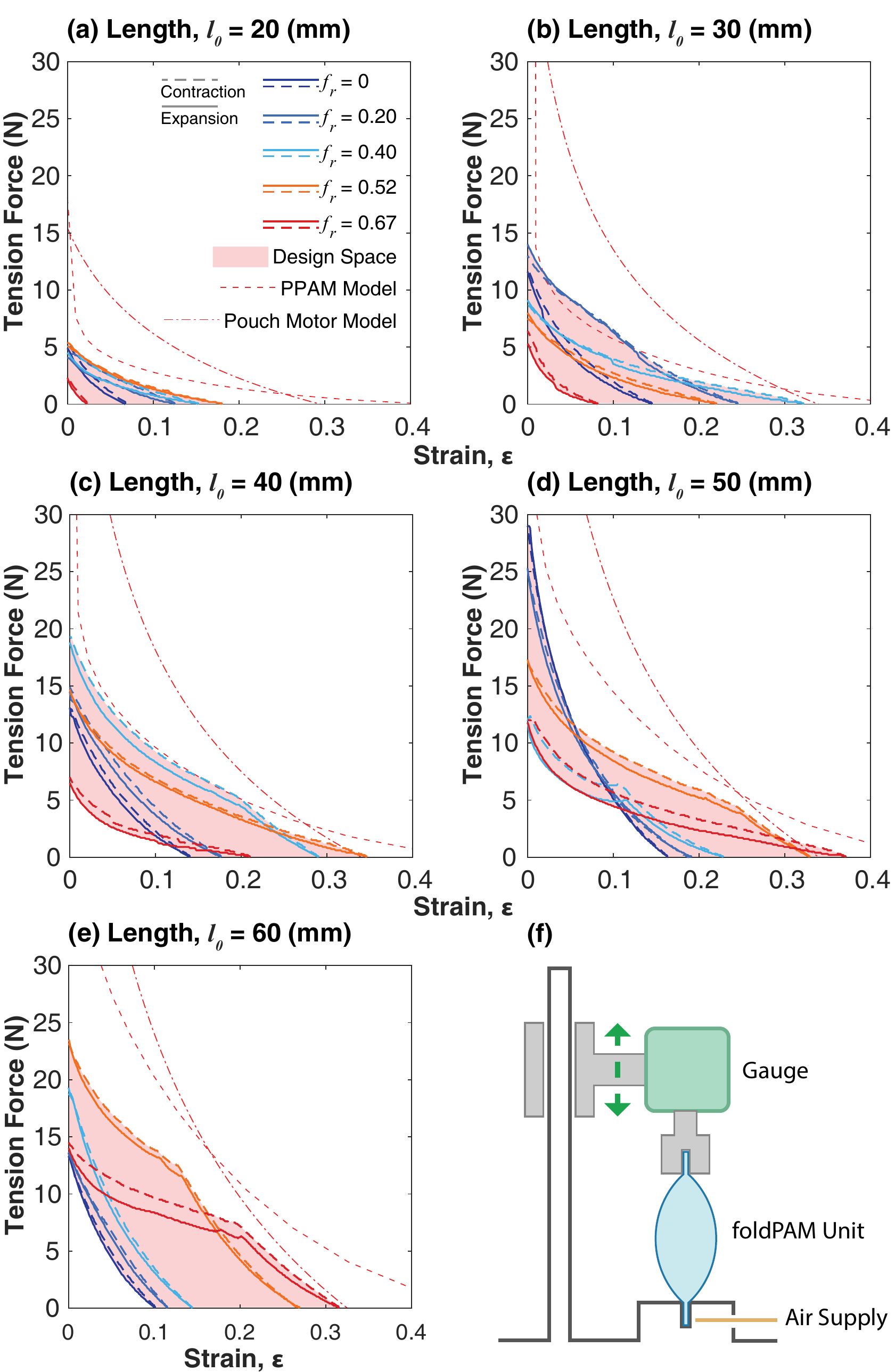}
    \caption{(a)-(e): force-strain plot of static foldPAM units. All units have a $W_0$ of 50mm and $l_0$ varies from 20-60mm, resulting in $a_r$ from 0.4 to 1.2. The PPAM and Pouch Motor models are superimposed with the data, and the design space bounded by the experimental force-strain curves is labeled by the shaded area. (f): Schematic of the test set-up.}
    \label{fig:static_force_strain}
    \vspace{-0.5em}
\end{figure}

\begin{figure}[tb]
    \centering
    \includegraphics[width=.95\columnwidth]{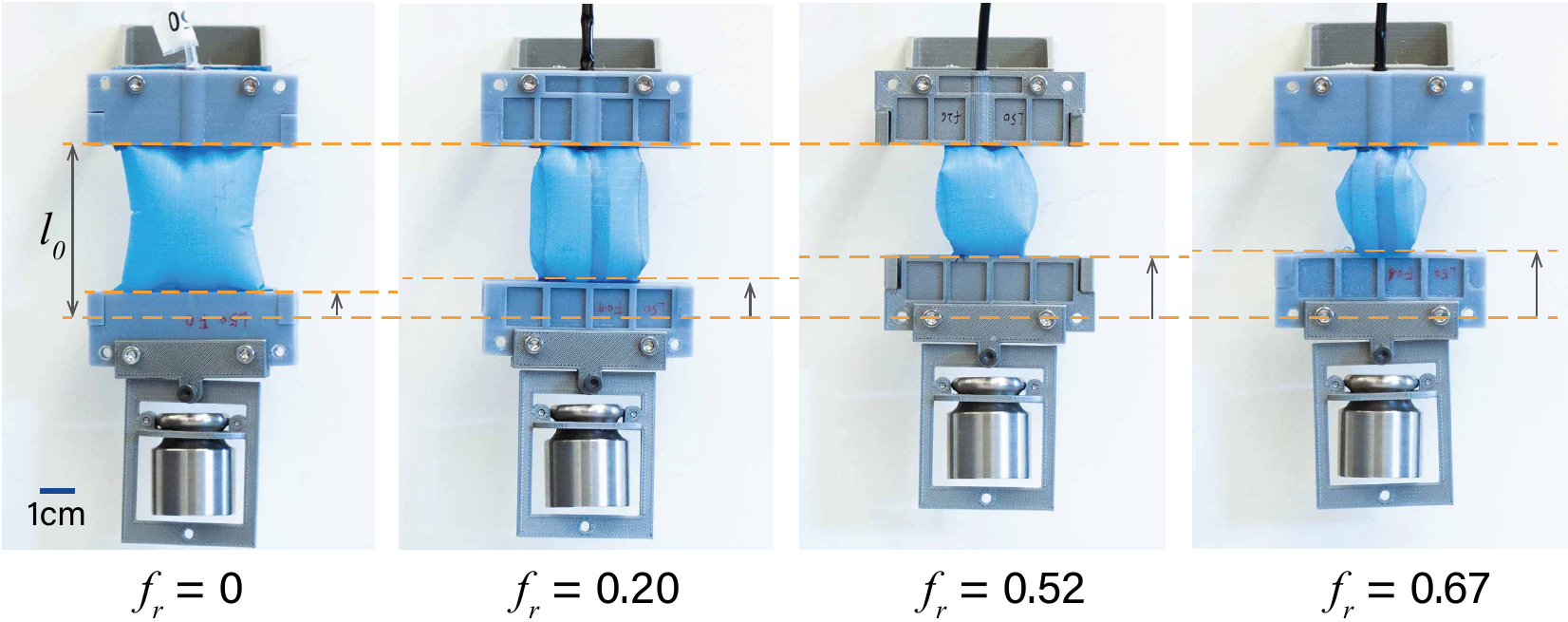}
    \caption{A comparison of static-geometry foldPAMs with fold ratios $f_r$ of 0, 0.20, 0.52, and 0.67. The units all have an initial length $l_0$ of 50mm, are inflated to a pressure of 12.4 kPa, and subjected to a load of 1N. Note that the amounts of strain observed here reflect their relative magnitude in Figure~\ref{fig:static_force_strain}(d).}
    \label{fig:static_demo}
    \vspace{-1.3em}
\end{figure}

\begin{figure}[t]
    \centering
    \includegraphics[width=.95\columnwidth]{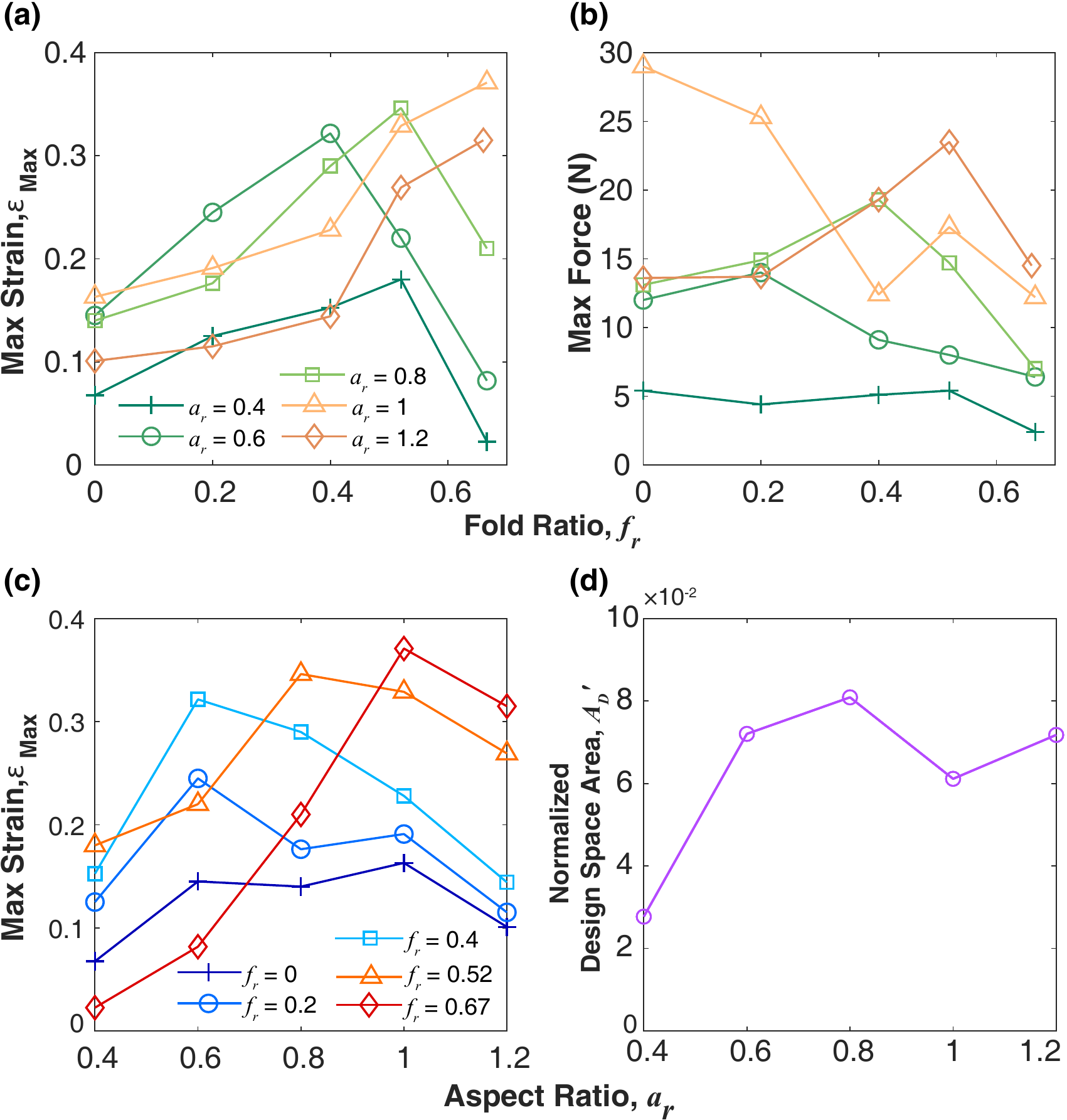}
    \caption{(a) The maximum strain of different $l_0$ over $f_r$’s. It is desirable to use the shorter length foldPAMs for their linearity, while longer foldPAM lengths can reach higher strains. (b) Maximum force of different $l_0$ over $f_r$’s. Maximum force decreases as $f_r$ increases. (c) The Maximum stress’ of different $l_o$ over five $a_r$’s. From $a_r$ 0.5 to 1, notice a flattening in $f_r$ 0 to 0.4 while we observe a steep increase in contraction ratio and decreasing at higher $f_r$ 0.52 and 0.67. (d) Area of Design Space and $a_r$ relationship which is monotonic positive. As $f_r$ increases, a higher max strain and lower force are observed.}
    \label{fig:sfs_analysis}
    \vspace{-1.5em}
\end{figure}
\subsection{Experimental Methods}
\label{sec:static_experiment}
We measured the force-strain relationships of a matrix of individual foldPAM units with different fold ratios and aspect ratios. The units were fabricated using the method described in Section \ref{stat_fab} with aspect ratios $a_r\in[0.4,1.2]$ with an incremental step of 0.2, and fold ratios $f_r = 0,\:0.2,\:0.4,\:0.52,\:0.67$. The unfolded width $W_0$ was held constant at 50mm for all units, so the $a_r$ gives initial lengths $l_0$ ranging from 20mm to 60mm. The testing units were sewn into a pair of custom rigid fixtures and mounted to a force gauge (Series 7, Mark-10 Corporation, NY, USA) on a motorized travelling test stand (ESM 303, Mark-10 Corporation, NY, USA). This setup is shown in Figure~\ref{fig:static_force_strain}(f). 

The force gauge was initially positioned such that the distance between the two ends of the tested unit was $l_0$ as defined in Figure~\ref{fig:fopam_paras}. After the foldPAM was pressurized to 12.4kPa, the force gauge travelled at a constant rate of 15mm/min, compressing the pressurized actuator to the zero-force, maximum contraction state, and then returning to the initial position. The force was recorded at a rate of 5Hz and the corresponding strain of the actuator was deduced from the known constant travel rate. The testing units pressure was regulated by a pressure control valve (QB3, Proportion Air, McCordsville, IN, USA) for all tests.

\vspace{-1em}

\subsection{Result and Discussion}
\subsubsection{Force-Strain Characteristics}
The results of the force-strain experiment can be seen in Figure~\ref{fig:static_force_strain}(a)-(e) broken down by length. We see that, in general, as $f_r$ increases the max strain increases and the max force decreases. For smaller fold ratios, i.e. $f_r=$0, 0.2, and 0.4, the max strain increases as $l_0$ or $a_r$ increases. Figure~\ref{fig:static_demo} shows foldPAMs with $l_0 = 50$mm inflated to a pressure of 12.4kPa and subjected to a constant load of 1N, showing a ``slice" of the data. Figure~\ref{fig:sfs_analysis}(a) and (b) show the max strain and max force as a function of fold ratio, and Figure~\ref{fig:sfs_analysis}(c) shows the max strain as a function of aspect ratio. \revision{The maximum strain obtained ranges from 0.02 to 0.37, indicating that we may program the actuator to match or approximate the maximum strain of certain existing actuators, such as fPAM ($\epsilon_{max} = 0.31$, \cite{naclerio20}), McKibben Muscle ($\epsilon_{max} = 0.39$, \cite{tondu12}), and pouch motor ($\epsilon_{max} = 0.34$, \cite{greer17}); the maximum strain also approaches that of sPAM ($\epsilon_{max} = 0.45$, \cite{greer17}) and PPAM ($\epsilon_{max} = 0.41$, \cite{daerden01}).}

While these trends describe the majority of the data, we see some interesting behaviors at high aspect ratio and fold ratio. In several experiments with $f_r \geq 0.4$, we observe a ``kink" in the force-strain curve, dividing the curve into two linear regions and giving a slower rate of increase in force output towards zero strain. This behavior indicates a sudden change in volume expansion during the eversion of the folded pouch, and fabricating the actuator with a more compliant material would likely reduce the extent of the behavior.


Examining the max strain as a function of aspect ratio (Figure~\ref{fig:sfs_analysis}(c)) we can see behavior similar to previous experiments on pouch motors and sPAMs. For pouch motors, Greer et al. \cite{greer17} demonstrated that, at aspect ratios below $1$, the maximum strain approaches the theoretical limit. We see similar behavior at lower fold ratios, seen in a flattening observed in $f_r$’s 0, 0.2, and 0.4. For sPAMs, Greer et al. shows the strain increases with aspect ratio up to a peak near $a_r = 1.5$. We can observe the transition to sPAM-like behavior for fold ratios 0.52 and 0.67, with the maximum strain at the optimal aspect ratio approaching, but remaining under, the values reported for sPAMs.

\subsubsection{Relating to Model Prediction}
We can additionally use the data to analyze the models presented in Section~\ref{sec:static_model}. When these models are plotted against the data in Figure~\ref{fig:static_force_strain}, we note that the sPAM/PPAM model generally predicts higher force and strain output, and the pouch motor model often predicts a higher force output but approaches the experimental results near the maximum strain. The deviation may indicate a loss of both force and strain due to material elasticity and the different geometry of the foldPAM during volume expansion. For pneumatic actuators the expected force is given by $dF = PdV$, where $P$ is the internal pressure and $dF$, $dV$ are differential changes in the force and volume of actuator. At higher strain where the actuator is nearly fully inflated, we see less volume loss and the ideal pouch motor model predicts the actuator behavior accurately for $w_f$ near the value given in Equation (\ref{eqn:wf_circ}), especially for greater actuator length and higher aspect ratio. However, at lower strain, the folded portion of the actuator reduces the volume of the actuator when not fully inflated, resulting in a lower force than the models predict. This effect is particularly pronounced at higher fold and aspect ratios. At maximum fold ratio, the folding limits the lateral expansion and results in a non-circular cross section, which deviates from the assumption by the sPAM/PPAM models, and the model approximates the measured data only for a limited number of cases. At low aspect ratios such as when $a_r \leq 0.8$, the apparent aspect ratio ($l_0/W$) of the actuator remains low even with a maximum fold ratio, making it more appropriate to consider the actuator as an under-inflated pouch motor than a sPAM/PPAM. Overall, while the pouch motor model partially predicts the force-strain relation at the ideal pouch motor model state, there lacks analytical models that predicts the behavior of the foldPAM near the other two states well without significant correction. Developing better insights to the behavior of foldPAM is a focus of ongoing work.

\subsubsection{Design Space Analysis}
The results in Figure~\ref{fig:static_force_strain} for each length of the tested unit bound an area in the force-strain plane. As we expect continuously changing behavior when the parameter $f_r$ is tuned, every combination of force and strain within the bounded area is reachable for some $f_r$. Thus, we define the bounded area as the ``design space", $A_D$, representing the amount of reachable area on the force-strain plane for the foldPAM. Clearly, as $A_D$ depends on the minimum and maximum force output, it would be a function of the internal pressure and the cross-section area of the actuator; and since it is difficult to measure the cross-section for different fold ratios, we define a normalized design space $A_D'$ based on pressure and the unfolded area of the actuator, which correlates with the cross-section area, as follows:
\begin{equation}
    A_D' = \frac{A_D}{a_r{W_0}^2P}.
\end{equation}
The value of $A_D'$ obtained at each aspect ratio is shown in Figure~\ref{fig:sfs_analysis}(d). We see that except for $a_r = 0.4$ where both force and strain diminishes, $A_D'$ fluctuates near a constant value of approximately 0.06. This indicates that the design space area is relatively independent of the aspect ratio of the actuator, and rather scales with pressure and dimension.

Overall, the experimental data suggest that the foldPAM behavior is similar to both pouch motor and sPAM behavior depending on aspect and fold ratio. The design space analysis suggest foldPAMs can be made for a wide range of force-strain curves, but additional modeling would be needed to a priori predict a foldPAM's behavior.

\section{Active Geometry foldPAM}
The previous section has shown that a range of force-strain relationships can be realized by actuators fabricated at different fold ratios. While this shows that we can pre-program an actuator's behavior with folding, it also suggests that we could also control an actuator by actively varying its fold ratio. This capability may be interpreted as: (1) a method for tuning the stiffness of the actuator for a given pressure; or (2) a way to control the motion of the actuator through geometry. In this section we implement a tunable fold ratio device and use the device to regulate its length at a desired position with feedback control.

\vspace{-1em}

\subsection{Design and Fabrication}
\begin{figure}[t]
    \centering
    \includegraphics[width=.92\columnwidth]{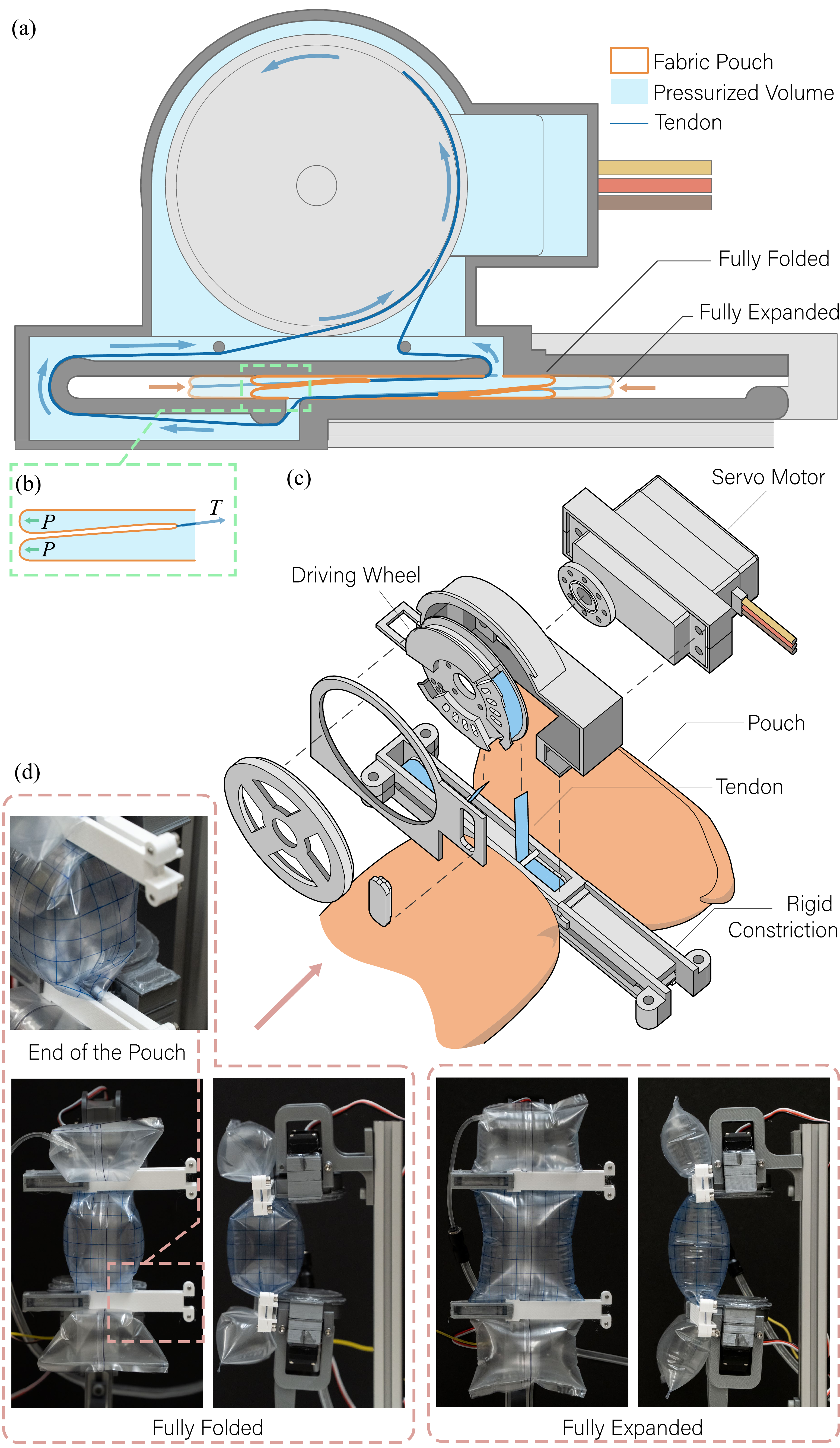}
    \caption{\revision{Illustrations of the Active foldPAM prototype. (a): a section view of the driving mechanism indicating the change in the folding of the pouch from the fully expanded ($f_r=0$) state to the fully folded state; (b): Enlarged view of the fabric pouch showing the antagonistic pressure force P and the tension force T in the tendon; (c): Exploded view of the rigid frame assembly on one end; (d): Photos of an Active foldPAM prototype with a transparent TPU pouch with lines to aid articulating the shape of the pouch.}}
    \label{fig:active_concept}
\vspace{-1.8em}
\end{figure}
As shown in Figure~\ref{fig:active_concept}, the device consists of a fabric pouch attached to a pair of rigid frames. The frame constricts the ends of the actuator to have a thin geometry as in the static foldPAM units and also allows actuation of the fold-ratio adjustment. The portion of the pouch bounded by the rigid frames produces the strain of the actuator; while the portions extended beyond the frames does not contribute to the contraction, they make deformation of the pouch easier. 

To realize a change in fold ratio, we need to enable linear motion of the points at the end of each edge of the actuator. This is enabled by an antagonistic, tendon-driven mechanism embedded in the rigid frames, as illustrated in Figure~\ref{fig:active_concept}. At each end of the actuator, a pair of active tendons is connected to the inside of the fabric pouch and routed through a channel in the rigid frame, with their other end attached to a wheel driven by a servo motor. The channel and the housing for the servo motor are sealed to retain pressure in the actuator, creating the pressurized volume shown shaded in Figure~\ref{fig:active_concept}(a). 
The edges of the fabric pouch are pulled inwards at equal rate when the wheel rotates counter-clockwise, increasing the fold ratio $f_r$ for the unit. When the wheel rotates clockwise, it releases the tendon, and the interior pressure in the fabric pouch causes it to expand laterally, thereby reducing the $f_r$.

In our implementation, the fabric pouch is fabricated with the same materials and methods as the static-geometry units. The structural components of the rigid frame are 3D-printed from PLA material. A thin layer of solvent-based, gap-filling adhesive (TAMIYA Inc., Shizuoka, Japan) is applied to the interior surface of all pressurized chambers to ensure a seal. A servo motor (FT5330M, Feetech, Guangdong, China) operating at 7.4V is installed on each of the rigid frames to actuate the wheels. The assembly of the actuation mechanism is shown in exploded view in Figure~\ref{fig:active_concept}(c). 

\revision{To illustrate the behavior of the proposed mechanism, we constructed a prototype with a pouch fabricated from transparent thermoplastic polyurethane (TPU) film. This allows the internal folding of the pouch to be seen. Figure~\ref{fig:active_concept}(d) gives a comparison of the inflated pouch at zero and maximum folding. The folding takes a similar shape as in a static unit of similar $a_r$ and $f_r$. The rigid constriction (at the fold) needs to have a finite gap to allow the pouch material to slide past itself smoothly, unlike the zero thickness of the sealed actuators. However, this discrepancy does not have a significant effect on the overall geometry of the actuator.} 

\vspace{-1em}

\subsection{Experimental Validation}
\begin{figure}[tb]
    \centering
    \includegraphics[width=.95\columnwidth]{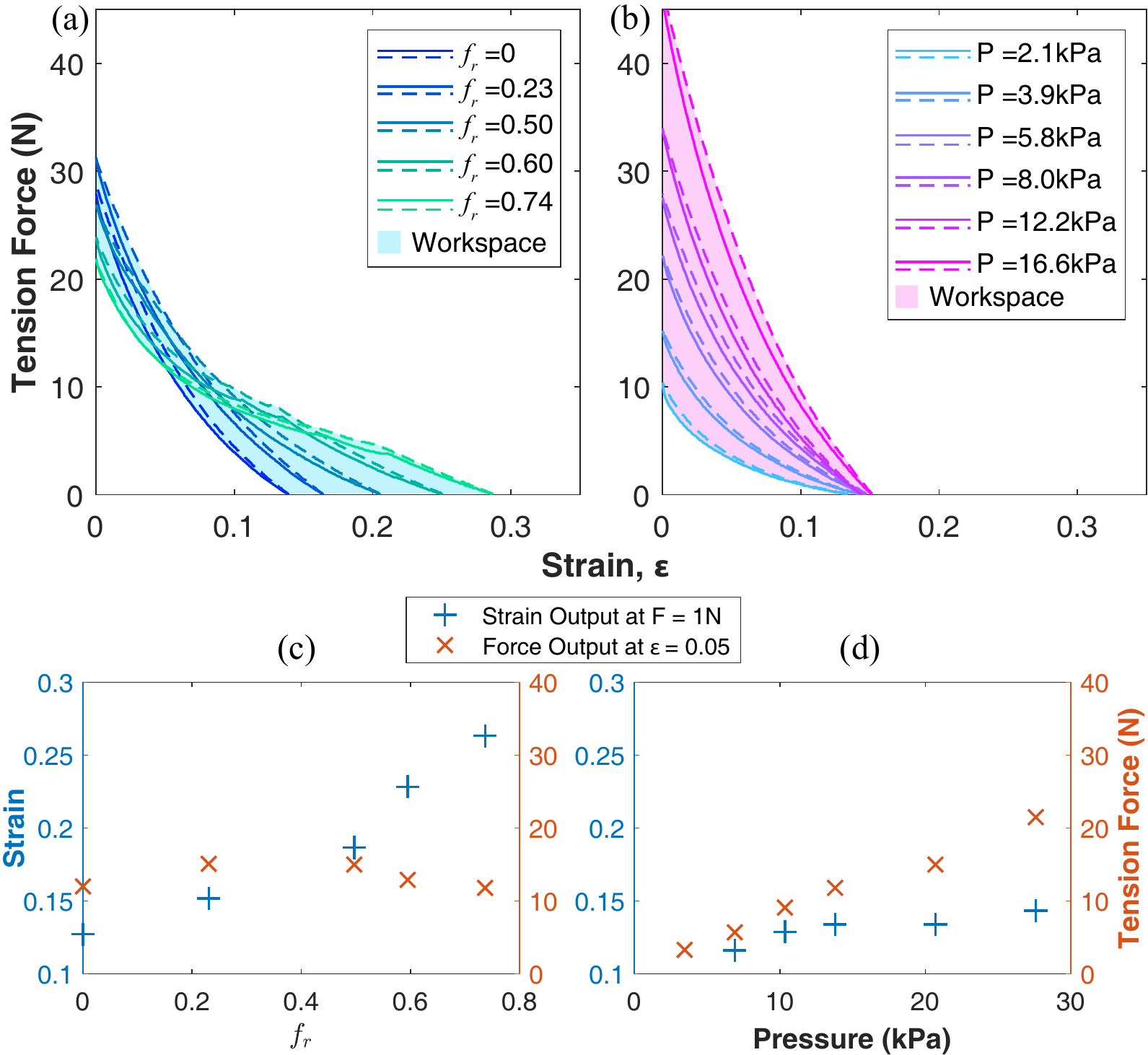}
    \caption{The force-strain relation of the Active foldPAM device, when (a) the internal pressure is kept at 8.0kPa and $f_r$ is varied between 0 and 0.74 and (b) under a constant $f_r = 0$ and the pressure changes varies from 2.1kPa to 16.6kPa. The area bounded by the force strain curves give the workspace of the Active foldPAM device.}
    \label{fig:active_1}
\vspace{-1.7em}
\end{figure}
\subsubsection{Force-Strain Characterization}
\label{sec:active_force_strain}
The experiment is performed with the same testing apparatus as in Section \ref{sec:static_experiment}, with modified fixtures on the test stand and the force gauge to install the Active foldPAM prototype. To compare the effects of actively controlling the geometry and the internal pressure of the device, we perform two sets of experiments, where the measured fold ratio $f_r$ varies from 0 to 0.74 at a constant pressure of 8.0kPa for the first experiment, and the internal pressure varies from 2.1kPa to 16.6kPa at a constant $f_r$ of 0 for the second. The upper bound of the tested values are the maximum folding and pressure that the prototype can sustain. Notably, the maximum measured $f_r$ exceeds the theoretical limit of the fold ratio at 0.67 due to material elasticity. The result of the experiment is shown in Figure~\ref{fig:active_1}.

While one can observe changes in both maximum strain and maximum tension as the fold ratio is varied (Figure~\ref{fig:active_1} (c)), only the maximum force shows monotonic change when the pressure is varied (Figure~\ref{fig:active_1} (d)), which is consistent with the previous literature \cite{niiyama15}. Except when $f_r$ is small, increasing $f_r$ results in decreased maximum force and increased strain (i.e., changing the fold ratio ``trades" tension for strain). Similar to the ``design space" for static foldPAM units, we define a ``workspace" for the Active foldPAM, which is the area bounded by all the curves on the force-strain plane and hence contains all the combinations of force and strain attainable for the device by varying the control input. 

The workspace shows distinct area distribution under the two control inputs tested: for the geometric control, the workspace covers a relatively large area at low force, whereas pressure control allows little adjustment at such force output as the force-strain curve converges to a single point. On the other hand, while the workspace under pressure control spans a large range of forces at low strain, the workspace for geometric control shrinks significantly at such strain and displays a non-linear fold-to-force mapping. For example, consider two cases of constant output: in Figure~\ref{fig:active_1}(c), with geometric control, the device allows its strain to vary across a range of approximately 0.3 while giving a constant 1N force output, while pressure control in Figure~\ref{fig:active_1}(d) only yields a range of 0.15; however, pressure control allows a larger range of force output under a constant $\epsilon$ of 0.05, as compared to the geometric control that effectively cannot tune the force.

\begin{figure}[tb]
    \centering
    \includegraphics[width=.88\columnwidth]{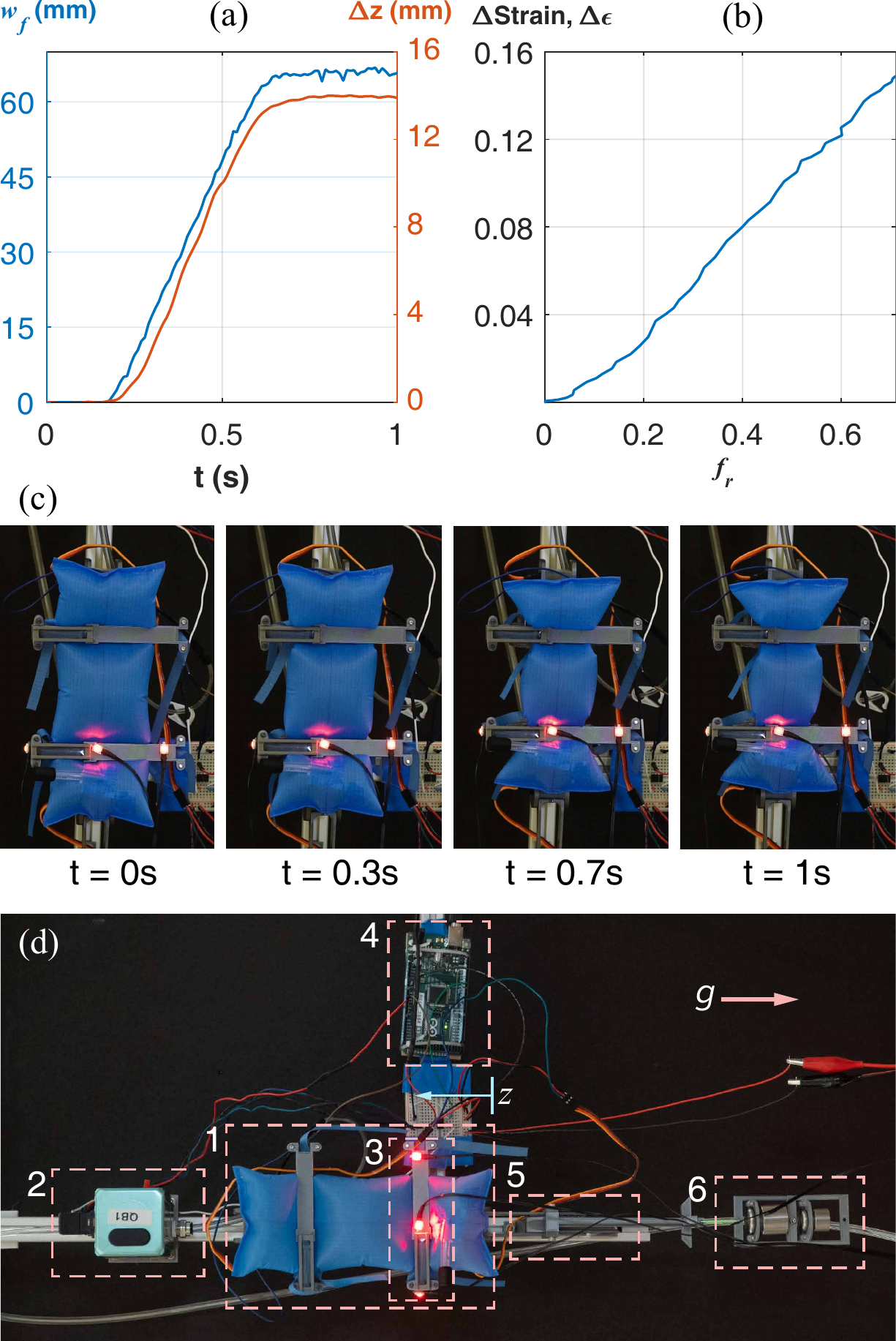}
    \caption{(a): open-loop response of a vertically placed Active foldPAM device, when the device is commanded to travel from 0 fold to the maximum folding possible. The plot shows the folded length $w_f$ estimated from servo angle feedback and the displacement at the lower end of the device. (b): The relation between fold ratio and the change in strain obtained from the result in (a). (c): The Active foldPAM during the experiment, from $t = 0s$ to $t = 1s$. (d): The experiment set up used in Section \ref{sec:active_ol} and \ref{sec:active_cl}, with the downward direction indicated by $g$ (1: The Active foldPAM prototype; 2: pressure control valve; 3: motion capture marker; 4: microcontroller; 5: guiding rail; 6: external load).}
    \label{fig:active_2}
    \vspace{-1.5em}
\end{figure}
\subsubsection{Open-Loop, Fold-Ratio-to-Output Characterization}
\label{sec:active_ol}
As shown in Figure~\ref{fig:active_2}(d), the prototype is mounted vertically with the lower end sliding freely in a guiding slot, which constrains the motion to be one-dimensional. The setup is placed in the view of a motion capture system (Impulse X2E, PhaseSpace Inc., CA, USA) and three position tracking markers are placed on the rigid frame at the lower end. The Active foldPAM is initially at zero folding with the driving motor set to $0\degree$ and, at $t = 0s$, the servo motors mounted on both ends begin rotating to a set point of $160\degree$, resulting in a maximum $w_f$ of 67mm. The device pressure is 8.0kPa, controlled by a pressure control valve (QB3, Proportion Air, McCordsville, IN, USA), and it is only subjected to its own weight, which is approximately 100 grams. Both the servo and the pressure control valve are interfaced to a computer with an Arduino Mega 2560 microcontroller, which also transmits angle feedback in real-time. The value of $w_f$ is estimated from the servo angle feedback with a fitted polynomial.

The recorded servo angle and position output of the experiment are presented in Figure~\ref{fig:active_2}(a), (b), and the motion sequence of the prototype is shown in Figure~\ref{fig:active_2}(c). While the servo motor starts to rotate at $t=0s$, the motion of the device starts at approximately 0.2s, indicating a backlash due to the fabric pouch and tendon elasticity. A total travel ($\Delta z$) of 14mm is observed. The value of $w_f$ and $\Delta z$ in Figure~\ref{fig:active_2} is normalized with respect to $W_0$ and $l_0$ respectively, giving the relation between fold ratio $f_r$ to change in strain $\Delta\epsilon$ in Figure~\ref{fig:active_2}(b), which appears to be highly linear.

\vspace{-1em}

\subsection{Closed Loop Control through Geometry Change}
\label{sec:active_cl}
\begin{figure}[tb]
    \centering
    \includegraphics[width=.9\columnwidth]{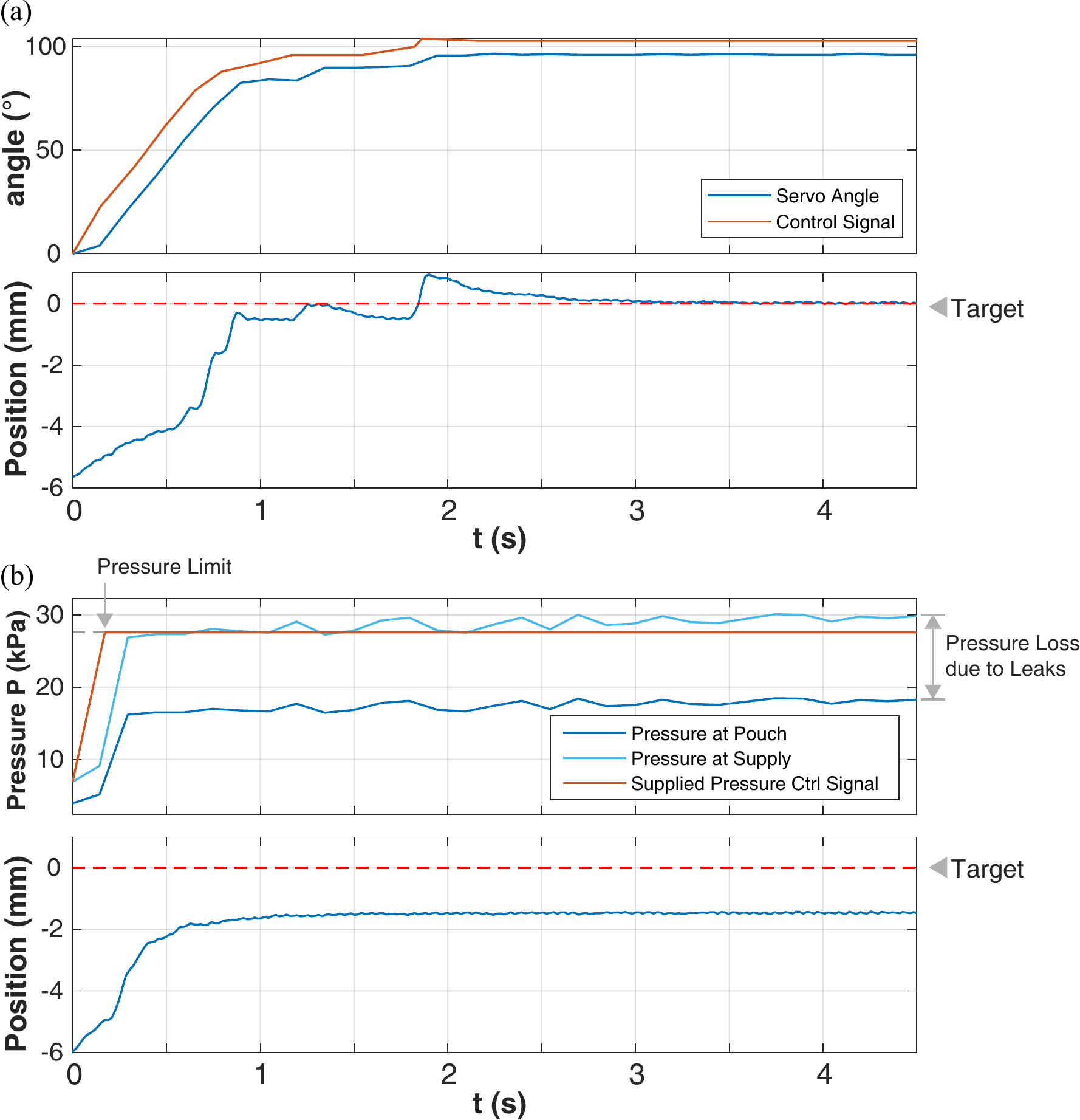}
    \caption{A PI control loop is implemented for the Active foldPAM device to regulate its position output when a step change in load of approximately 150 grams is applied. The figure shows the response of (a) geometry-based control and (b) pressure-based control. For (b) both the pressure feedback value from the supply and the actual pressure in the prototype is shown.}
    \label{fig:active_3}
    \vspace{-1.5em}
\end{figure}
Finally, we demonstrate that an Active foldPAM device can realize closed-loop control of position output. In particular, we suppose a scenario in which a step change load of 150g mass (1.47N under gravity) is applied to the initially unloaded foldPAM and the device attempts to restore its position by adjusting the fold ratio, while keeping a constant internal pressure of 3.9kPa. As a comparison, a separate experiment with a constant fold ratio of 0 uses the pressure to compensate for the displacement due to the applied load. The pressure supply is commanded to vary between 6.9kPa and 27.6kPa; \revision{the pressure at the pouch is controlled indirectly with feedback at the supply, and achieves a lower range of 3.9kPa to 16.7kPa due to leaks in the prototype}. The same set up for the open-loop displacement characterization is used in this demonstration. The real-time position feedback from the motion capture system is used to compute the error value, and a proportional-integral (PI) controller is used to generate the control signal for the servo motor and the pressure regulator for the two experiments respectively. The control loop runs at a frequency of 6 Hz.

Figure~\ref{fig:active_3} shows the results. The two experiments see an average initial error of 5.8mm. Due to the characteristics of the controller, in Figure~\ref{fig:active_3}(a) the control signal for the servo motor gradually increase, whereas the pressure control value in Figure~\ref{fig:active_3}(b) jumps instantaneously to the saturating level. At steady-state, the geometric control results in near-zero error, while a 1.5mm error is observed for pressure control. This reflects the observation in Section \ref{sec:active_force_strain}, that geometric actuation results in a larger strain workspace at a small, constant load, while the ability of the pressure to compensate for strain errors is limited, \revision{and in this case requires a pressure that is infeasible due to hardware constraints. It should be acknowledged that the performance of pressure control is limited by the prototype's ability to operate at sufficiently high pressure, and does not necessarily suggest that pressure control is always unable to achieve zero error. However, this does suggest that geometry actuation has higher robustness against leaks in the system.}

\section{Conclusion and Future Work}
In this paper, we presented a new design of a pneumatic actuator for which the force and strain can be tuned by varying the end geometry, namely the width of the laterally folded portion. This feature can be used for both pre-programming and actively controlling an actuator unit. With this principle, we explored the concept of ``geometric control", as opposed to the traditional pressure-based control strategy, and showed that it responds linearly to the control input and enables a different workspace than pressure control alone.

Future work on foldPAMs will focus on modeling the actuator behavior, for which approximations using existing models gives limited accuracy. For Active foldPAM units, the actuation of the geometry is currently bulky so a compact, lightweight source of actuation is needed to implement geometric change at a form factor that is favorable for applications. Finally, the concept of geometric control raises a wide range of new questions to explore, such as its applicability to other types of pneumatic actuators, and the potential of achieving programmable actuator trajectory with passive geometry-changing mechanisms.






\balance
\bibliographystyle{IEEEtran}
\bibliography{myReferences}

\end{document}